# Mechanomyography based closed-loop Functional Electrical Stimulation cycling system

Billy Woods, Mahendran Subramanian, Ali Shafti & A. Aldo Faisal, *Members, IEEE*

*Abstract*—**Functional Electrical Stimulation (FES) systems are successful in restoring motor function and supporting paralyzed users. Commercially available FES products are open loop, meaning that the system is unable to adapt to changing conditions with the user and their muscles which results in muscle fatigue and poor stimulation protocols. This is because it is difficult to close the loop between stimulation and monitoring of muscle contraction using adaptive stimulation. FES causes electrical artefacts which make it challenging to monitor muscle contractions with traditional methods such as electromyography (EMG). We look to overcome this limitation by combining FES with novel mechanomyographic (MMG) sensors to be able to monitor muscle activity during stimulation in real time. To provide a meaningful task we built an FES cycling rig with a software interface that enabled us to perform adaptive recording and stimulation, and then combine this with sensors to record forces applied to the pedals using force sensitive resistors (FSRs); crank angle position using a magnetic incremental encoder and inputs from the user using switches and a potentiometer. We illustrated this with a closed-loop stimulation algorithm that used the inputs from the sensors to control the output of a programmable RehaStim 1 FES stimulator (Hasomed) in real-time. This recumbent bicycle rig was used as a testing platform for FES cycling. The algorithm was designed to respond to a change in requested speed (RPM) from the user and change the stimulation power (% of maximum current mA) until this speed was achieved and then maintain it.**

## I. Introduction

Functional Electrical Stimulation (FES) systems enable muscles to be contracted even though the neurological path to the brain through the spine has been broken [1]. Although the primary intent of FES evoked exercise is to restore mobility in paralysed muscles, it can also increase health and life expectancy of Spinal Cord Injury (SCI) patients [2-4]. A high rate of muscle fatigue and Autonomic Dysreflexia are some of the limiting factors in the effective use of FES [5, 6]. But these limitations can be alleviated with more efficient control methods.

Raw electromyography (EMG) signals can be used to measure the force of contracting muscles. Previously, [7] EMG signals were used as a feedback control protocol for an FES cycling system. However, limitations of EMG sensors include precise placement, sensitive response to impedance changes in the skin caused by sweating and incorrect measurements caused by FES induced electrical artefacts [9]. Mechanomyography (MMG) is a promising alternative that is not affected by the limitations of EMG, which makes it advantageous for FES cycling applications [8].

A proportional-integral-derivative (PID) controller is a closed-loop feedback mechanism that can continuously monitor an error value that is calculated from the difference between the desired set point and a measured variable [10]. PID controllers have previously been used in FES applications. A PID controller was developed and used to control muscle stimulation pulse width to drive an FES assisted walking gait aided by a Spring Brake Orthosis (SBO). The results of this study showed that PID control was effective at controlling the extension of the leg during the use of the SBO during walking [11]. Another example of a PID controller being used in closed-loop FES applications is a positional control of wrist movements for patients with quadriplegia at C3 or C7 levels of SCI. The effective design of this PID controller to monitor the error signal between the desired position of the hand and the actual position, enabled stable and accurate movements to be carried out [12]. The PID controller is deemed to be a simple, reliable and well-established control method that has been applied to many FES systems. However, PID controllers are designed to be applied to linear time-invariant systems, whereas FES applications are non-linear and time-varying. This means that PID controllers are subject to parameter variation and exogenous disturbances, and therefore may not be the best choice for real world applications [13].

Many promising results have been obtained from studies into the efficiency of FES cycling, but most have focused around theoretical or modelled systems. A stationary rider and a cycle can be thought of and modelled as a closed chain mechanism. A biarticular muscle model was developed by [14] to test the efficiency of a closed loop FES cycling system that tracked the cadence and velocity of a stationary rider. This system was then tested on seven healthy volunteers and one with a diagnosis of Parkinson's Disease. The pulse width of the FES stimulation was altered in real time and would adapt with the aim of making the rider cycle as close to the proposed closed chain model as possible.

* This project contributed to the Team Imperial effort at the Cybathlon 2016 (Klothen, CH). The authors would like to acknowledge the advice and support of Paul Moore (Active Linx, Peterborough, UK) and Rik Berkelmans (Berkel Bikes, Sint-Michielsgestel, NL) MS was supported by an EPSRC DTA. AAF was supported by eNHANCE (http://www.enhance-motion.eu) under European Union's Horizon 2020 research and innovation programme grant agreement No 644000.

B. W., M. S., A. S., and A.A.F. are with the Brain & Behaviour Lab, Dept. of Bioengineering & Dept. of Computing, Imperial College London, South Kensington, SW7 2AZ, UK (Address for correspondence a.faisal@imperial.ac.uk).

From the experimental results this method could realize FES cycling close to voluntary tracking [14].

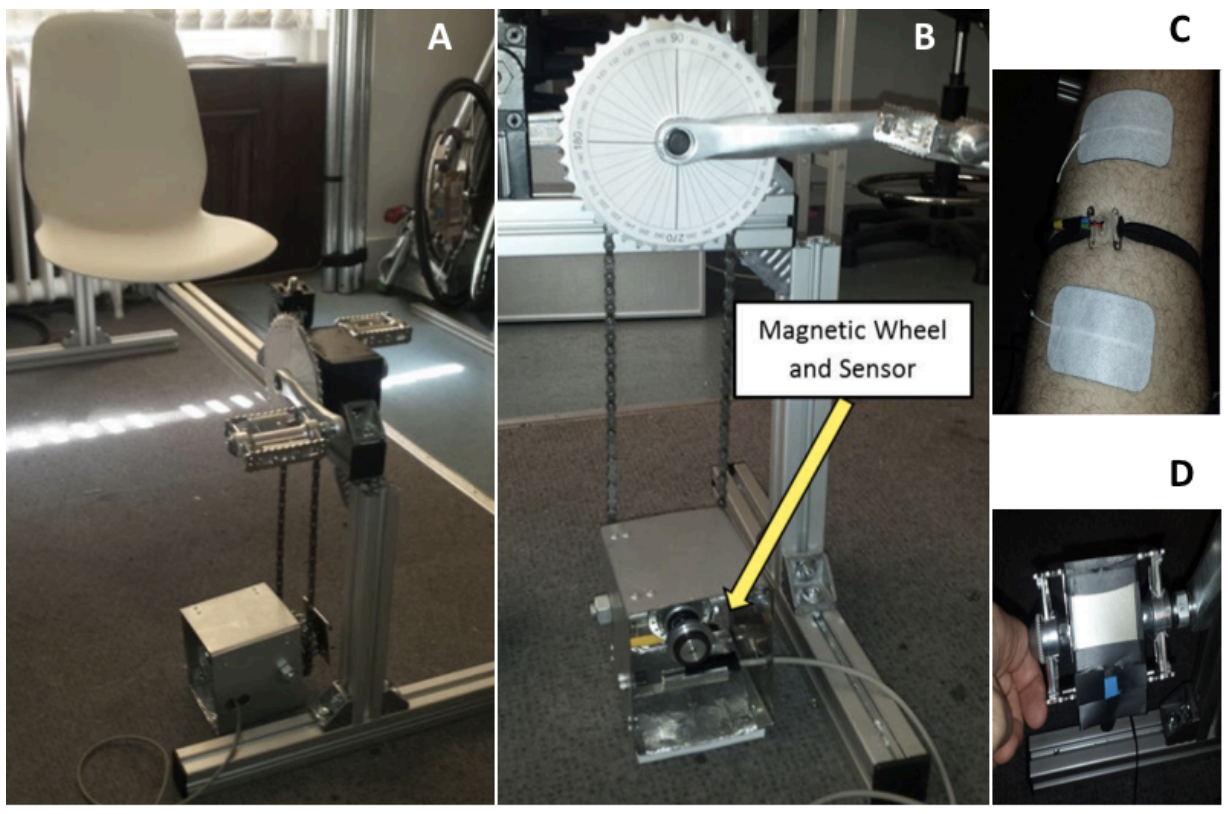

Figure 1. [A] Recumbent bicycle rig and [B] the crank angle encoder box with the magnetic wheel and sensor mounted. [C] Arrangement of FES electrodes (white pads) and MMG probe (transparent plastic held by black elastic band) and the Velcro strapping attachment of a force sensitive resistor. [D] Pedal sensors.

The primary focus of our work was to achieve closed-loop "on the fly" optimization for an FES system using pedal force and MMG sensors, giving the pilot control over the speed of cycling. The system must be able to make changes in the stimulation parameters based on feedback, until the desired speed has been reached, and then to maintain it. A recumbent cycling rig was constructed. This rig consisted of a frame on to which a seat was mounted. The seat could be moved forward and backwards, and a pedal crank was positioned at the front of the frame onto which two pedals were mounted. An off-axis crank box was connected to the pedal crank via a bicycle chain which enabled the two cranks to rotate together and featured a magnetic sensor that could detect movement in the crank. This rig and crank box setup can be seen in Fig. 1 and was used as a testing platform for the developed closed-loop system.

## II. Methods

### *A. Overview of the proposed closed-loop FES system*

A schematic layout of the closed-loop FES cycling system with all the proposed design elements can be seen in Fig. 2. USB-6009 Digital Acquisition Devices - DAQs (National Instruments, Berkshire, UK) were required to incorporate 2 pedal force sensors and 4 MMG probes. The interface board (I/F) and mechanomyography (MMG) board were created to house all the components necessary for signal processing.

The FES electrodes placement is as follows; two are placed on the quadriceps femoris, two on the hamstrings (one below the gluteal fold) and two on the gluteal maximus. The electrodes are placed so that as much of the muscle tissue is stimulated as possible. The MMG probes are placed central between the FES electrodes for each muscle group. The MMG probes were not used for the gluteal muscles as there was a danger that they could injure the skin due to the pressure on the seat.

Previous studies have shown that for FES cycling the quadriceps and hamstrings produce the most the power. The gluteal muscles were included in this study because stimulation of this area has been shown to reduce the possibility of pressure sores in SCI patients.

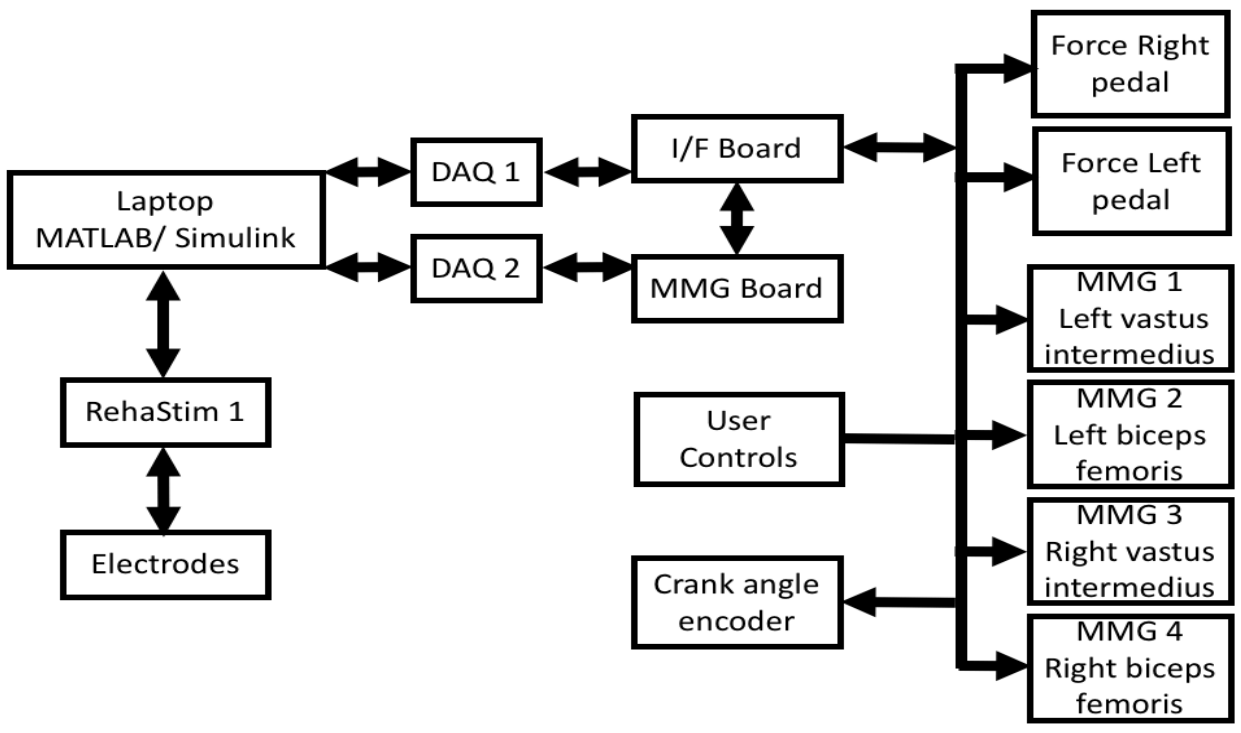

Figure 2. Schematic layout of the key elements of the proposed FES cycling system.

### *B. Development of the Electronics*

The electronics used in the FES cycling system were designed as an interface between data collected from the sensing elements and the data acquisition devices.

A novel design for the implementation of an ultra-low-cost MMG sensor was presented by [15]. This design utilized a low-cost condenser microphone that could detect lateral oscillations in the muscle fibers during contractions. This design was developed further. Two of these microphone transducers were housed in a specially designed chamber, one faces the muscle, and the other is used to detect background noise. One microphone will be placed in the center of the housing and detect the muscle contractions while the other is placed on the outside to detect acoustic background noise. By utilizing the two microphones together a noise cancellation effect can be achieved. The system is waterproof to reduce the risk of electrical contact with the skin during use. To provide the amplification for the output signal of the microphone an AD623 (Analog Devices, Massachusetts, USA) instrumentation amplifier is used. Each MMG probe requires two transducers and therefore two analogue inputs; each DAQ can, thus, process inputs from 4 of these constructed probes.

Two identical sensors were required to measure the forces applied by the left and right legs onto the respective pedals during cycling. Force Sensing Resistors (Interlink Electronics, California, USA) were chosen for this application. A simple bridge circuit otherwise known as a voltage divider was designed to convert forces applied on to the FSR into a voltage that could be read by the DAQ.

A user control was initially designed to allow the system to be switched on and off and control how much current would be applied to the muscles, this was achieved using a rocker switch and a potentiometer. This same control was also used for the closed loop control, the potentiometer was used to allow the user to control the desired output of the system or set-point (speed, RPM). Cycling is conducted through a full 360-degree motion, and the muscle groups used: quadriceps, hamstrings and gluteals, would only be stimulated through parts of that full revolution. To determine these angles the muscle groups had to be stimulated whilst observing the efficiency of the cycling motion. However, this method was cumbersome and time-consuming. A logic circuit was designed to adjust the stimulation current (mA) at the computer, for this stage of the experiment. Once these stimulation angles were found the same control interface could be used to switch the system on and off but also control the desired speed (RPM). The muscle stimulation strategy was to stimulate the muscle groups (quadriceps, hamstrings and gluteals) in accordance with the angles found during testing.

For the Incremental Encoder Interface, a magnetic sensor-MDFK 08 was used (Baumer Electric, Frauenfeld, CH). The outputs from the sensor are digital and the rising and falling edges of the square wave can be counted to calculate the change in angle. The pulse that indicates the zero position of the wheel was critical for use a crank angle measurement device. The SN74121 was used in a Monostable multivibrator circuit to increase the pulse width of the zero signal, making it easier to detect.

*C. Development of the Simulink control model*

Simulink was chosen to develop the embedded control system (Fig. 3). Simulink can be used to visually model a mathematical system using 'blocks' of code and an initial interface block for Simulink was provided with the RehaStim on their Science Stim website [16]. This interface block was designed to communicate with the RehaStim (Hasomed, Magdeburg, Germany) in binary code via a virtual COM port.

Figure 3. Simulink model that uses the digital input from the push button to control when FES stimulation is applied.

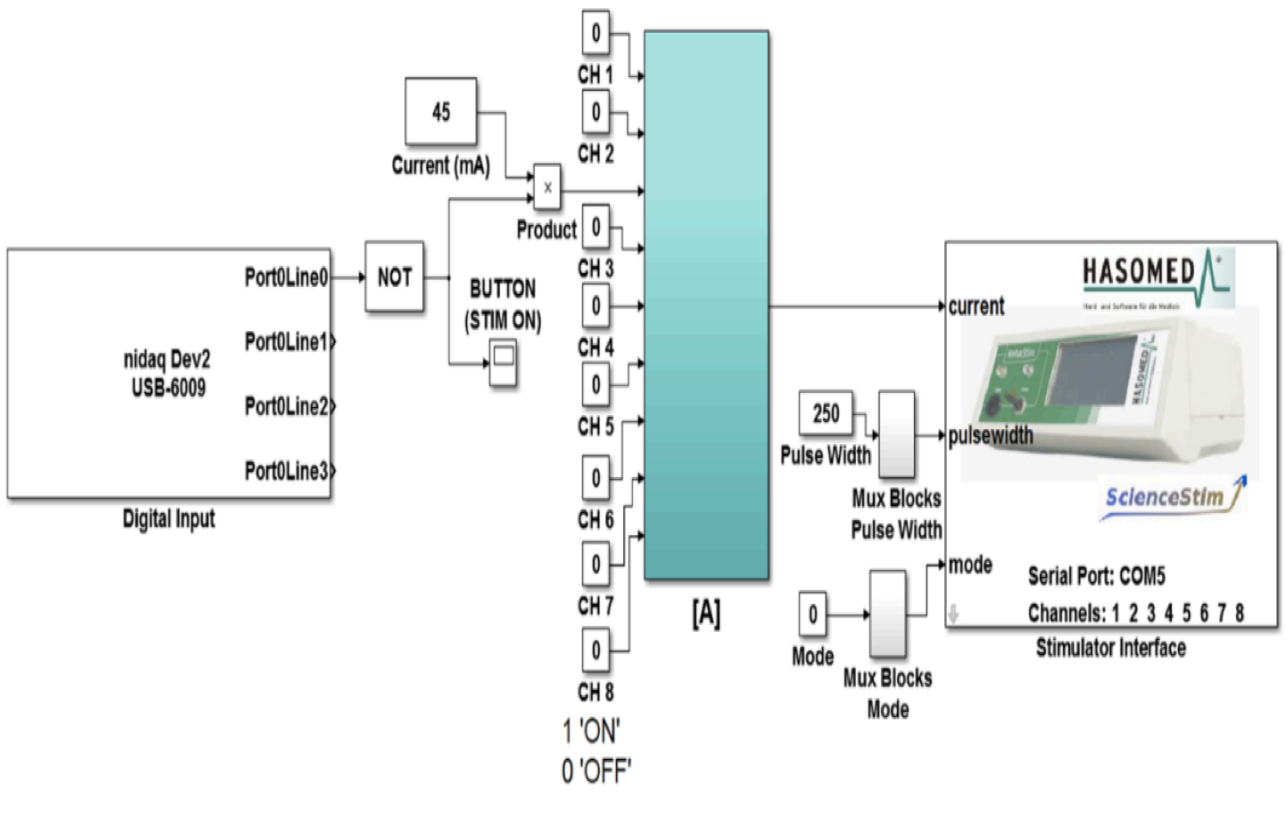

- *RehaStim 1 – Science Mode*

Simulink has many standard code blocks that can be used to construct models. One such block was the multiplexer (Mux) block. This block takes several single inputs and converts them into one vector output. By using Mux blocks and constant input blocks it was possible to create a basic working model for the stimulator. This model successfully produced output at the electrodes which were connect to the ReahStim. The next step at this stage was to improve this model so that the parameters could be externally influenced during a simulation. Product blocks were used in conjunction to the current input, which allow the values to be multiplied together. With these additional blocks in place it was possible to add user interface controls to the model. Toggle switches were added to the constant values attached to the input currents of each individual channel. Variable dials were attached to values associated with the stimulation current (mA), pulse width ($\mu$s) so that these could be changed during stimulation.

- *Designing the closed-loop algorithm*

To realize a system where the muscles would receive stimulation based on the crank angle a model had to be developed (Fig. 4). This model would take an input number that represented the current angle in degrees and would then switch the FES channels on or off based on the angles found during testing. By connecting all these elements, it was possible control the stimulation applied to the muscles based solely on the calculated angle of the crank. The next step in development was to integrate the user input. The rocker switch was used to provide the user a way of starting and stopping the stimulation.

Simulink had a PID control block in its library but this had a hybrid sampling frequency set up which was unusable with the data acquisition devices, which is why standard blocks were used to recreate the same effect.

Figure 4. Schematic layout for the closed-loop algorithm. Block [A] Takes input from the cycling rig to determine the current cycling speed. Block [B] Will take the difference in the requested speed by the user and compare it with the actual speed. A zero pulse will reset this calculation each time. Block [C] processes the logic behind how much stimulation is required at each muscle, and block [D] smoothes the output to avoid any sharp changes in stimulation. The system took several zero pulses before the angle measurement became accurate, so to avoid incorrect muscle contractions the system would not apply stimulation until an average had been found.

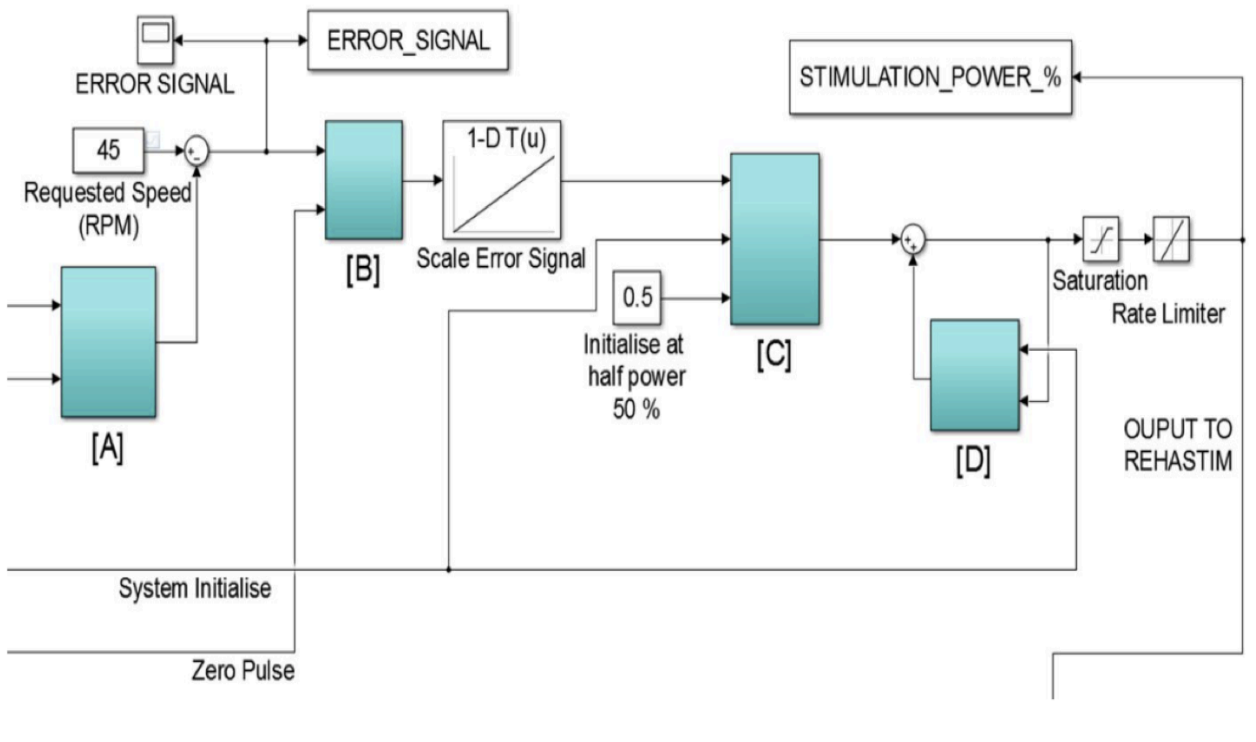

*D. Evaluating the performance of the crank angle encoder*

During initial testing with the Simulink models, it was observed that the number of pulses seen during rotations of the crank were not consistent. For the first stage of the evaluation, data was collected during simulation using the crank angle encoder model, with no stimulation to the muscles. The second part of the crank angle encoder evaluation involved investigating the system performance with active stimulation. Like the collection of data obtained from the previous section, except this time all pedaling motion is coming from the FES stimulated muscles. Three crank angle encoder models were created, one which takes the count from the previous revolution, one that takes the average of the last 2 counts and one that takes the average of the last 3 counts, to determine which model would be the most accurate. All three of these models were tested for 100, 500, 1000, 2000, 3000 and 6000 Hz sampling frequencies.

The 0 angle for the crank angle encoder was calibrated to show when the right pedal crank arm was pointing directly forward.

### *E. Evaluating the closed-loop system response*

The performance of the closed-loop system will be determined by how well the output stabilizes on the input specified by the user. The system has been designed so that the user can set a speed using the analogue input and the model will attempt to converge on to that speed in as few revolutions as possible. The crank angle encoder that averages the last three count values is used for this evaluation because it performed the best. A sampling speed of 100 Hz is also used because of the efficiency of performance and the ability to still use all the analogue inputs in real-time. The speeds (RPM) that will be tested are: 30, 40, 50, 60, 70, 80, 90 and 100.

## III. RESULTS

### *A. Calibration of the pedal force sensors*

The FSRs that were attached to each of the pedals were calibrated using cast iron weights in steps of 5 Newton's.

Figure 5. The calibration curves for the left and right pedal force sensing resistors.

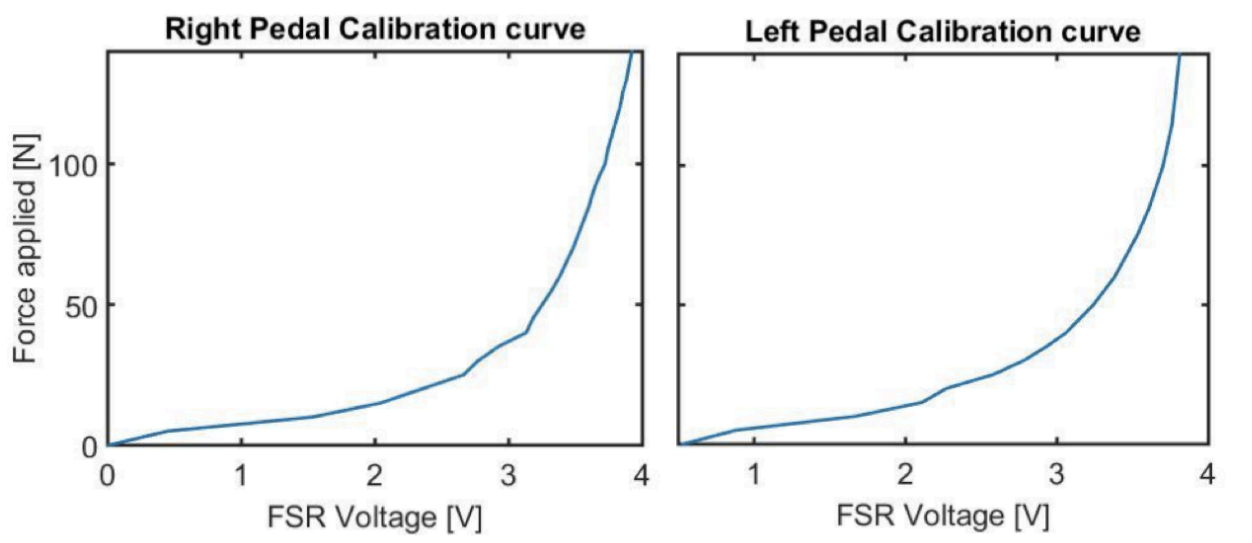

This data was entered into the relevant 1D lookup tables of the Simulink models for use with real-time signal conversion. The graphs of this data for both the left and right pedal FSRs can be seen in Fig. 5.

### *B. Manual Stimulation parameters*

The first stimulation parameters that were obtained related to the ideal stimulation current and pulse width for each muscle group. Tab. 1 shows these parameters. These values were then used in the Simulink model to obtain the stimulation angles that best activated each of the muscle groups. These parameters were obtained using a single healthy subject, and were the limit at which the stimulation was tolerable.

The current limit (mA) for each muscle group in Tab. 1 are the 100% stimulation power limits used during the testing phases. Throughout the experiment an FES muscle stimulation frequency of 40 Hz was used, this was not altered. During this testing phase it was also noted at which angle each of these muscle groups should be stimulated.

TABLE 1. STIMULATION PARAMETERS OBTAINED USING A SINGLE PUSH BUTTON SWITCH WHILST IN A RELAXED STANDING POSITION. THE PULSE WIDTH DEFINES THE TIME (s) OF THE BIPHASIC PULSES

| **Muscle Group** | **Current (mA)** | **Pulse Width (μs)** |
|---|---|---|
| Left Quadriceps | 40 | 200 |
| Left Hamstring | 50 | 225 |
| Left Gluteal | 45 | 250 |
| Right Quadriceps | 40 | 200 |
| Right Hamstring | 55 | 220 |
| Right Gluteal | 50 | 250 |

### *C. Evaluating the performance of the crank angle encoder*

The first stage of the crank angle encoder evaluation was the pulse counts and zero pulses. Data was collected focusing on only the crank counts at changing speeds for various sampling frequencies. The data were sorted according to the number of time samples acquired in between zero pulses.

Figure 6. The measured crank counts against time samples taken for six sampling frequencies. The red line shows the theoretical crank count that should have been detected at each revolution, regardless of time taken for the full revolution.

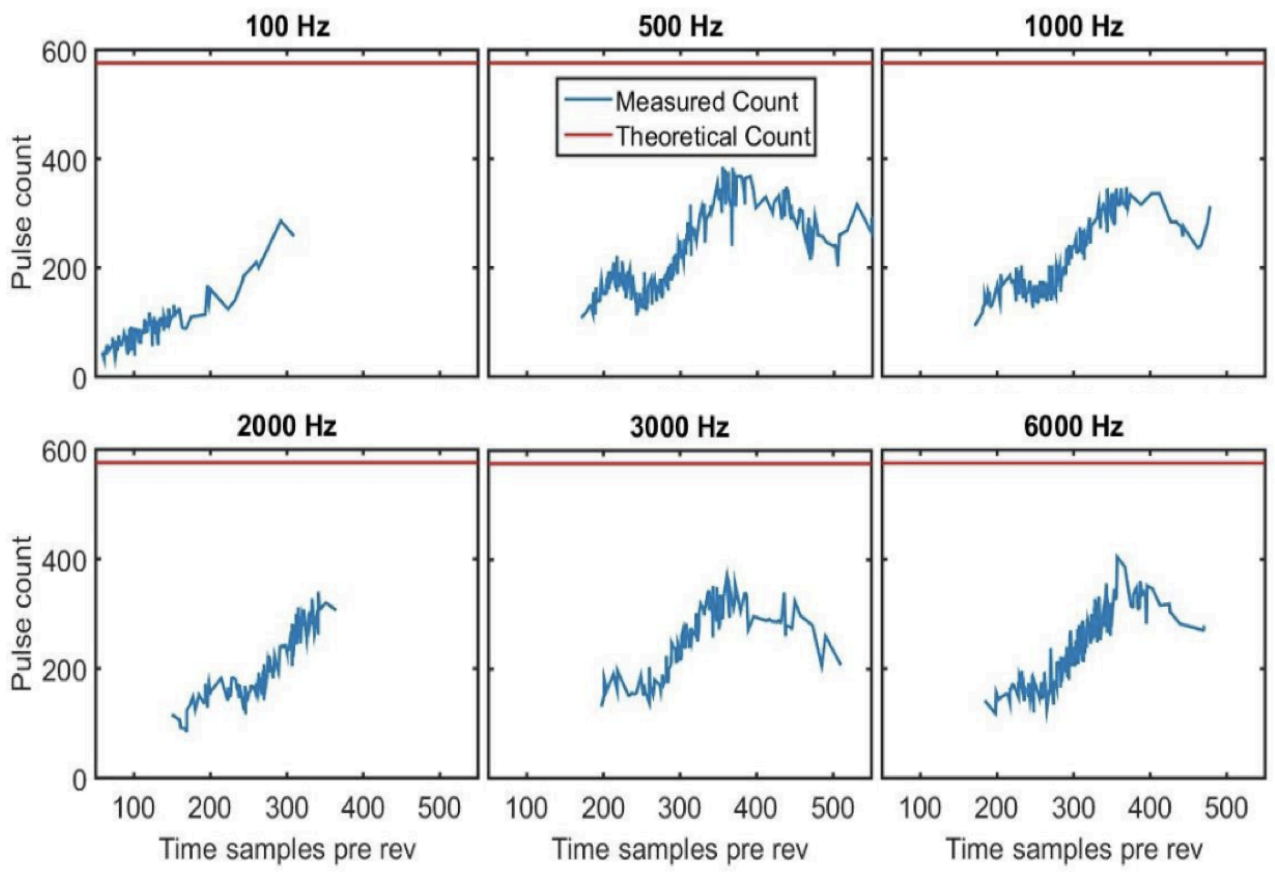

This equates to sorting the pulse counts for how much time had elapsed when the highest count had been achieved for each revolution, at which time the zero pulse would have reset this count. The red line in Fig. 6 indicates how many counts there should have been for each revolution, and it is clear from these diagrams that the counts are well below what is expected in all the collected data. There also does not appear to be an increase in the quality of the collected data as the sampling frequency is increased. Results show that at all sampling frequencies that the software was not able to record all the data.

TABLE 2. REAL-TIME DURATIONS FOR SIMULATIONS IN COMPARISON WITH THE USER SET TIME.

| Sampling Frequency (Hz) | Theoretical time between samples (s) | Time set for simulation (s) | Actual simulation time (s) [profiler] |
|---|---|---|---|
| 100 | 0.001 | 200 | 114 |
| 500 | 0.002 | 60 | 122 |
| 1000 | 0.001 | 30 | 122 |
| 2000 | 0.0005 | 15 | 123 |
| 3000 | 0.0003333 | 9 | 117 |
| 6000 | 0.000166666 | 5 | 106 |

The data in Tab. 2 shows how real-time is affected during the simulations ran in Simulink. At lower frequencies, the actual time will be faster than the number of seconds specified, whereas when the sampling frequency is 500 Hz or higher the real-time for simulations increases exponentially. This time dilation effect was likely the cause of pulse count errors

### *D. Evaluating the performance of the closed-loop system response*

For the evaluation of the closed-loop system response the last 3 average crank angle encoder model was used at a sampling frequency of 100 Hz. This low frequency meant that all the analog inputs for the entire system could be used in real-time. At a set speed of 40 RPM the error signal converged and oscillated around 0 at a stimulation power of around 60 %. For a set speed of 60 RPM the error signal still converged but 100 % was required to achieve this (Fig. 7). A set speed of 70 RPM did not converge. At this speed the system had ramped up the stimulation power to 100% within the first few iterations and then would stay there with the error value never reaching zero. Set speeds of less than 40 RPM were also not able to converge on a zero error because the stimulation value (%) never became strong enough to drive the legs to power the pedals round. A set speed of 50 RPM had a final stimulation of around 80%. Only set speeds of 40 and 60 RPM are shown here for clarity on the graphs.

## IV. Discussion

The final FES cycling system had a closed loop response and would adjust the stimulation parameters in real time to reduce the calculated error to a zero value. However due to inaccuracies of the crank angle data, the system never settles on a value of 0, rather oscillating around it. The elements of the closed loop system, including the user controls, biofeedback sensors and crank angle encoder were all tested using an oscilloscope and were shown to operate correctly. The inability of the whole system to truly reach and settle on a zero-error value therefore lies with the software elements.

To avoid any electrical interference from the FES, MMG sensors were used to measure activity in the muscles, which is a better option when used with FES in a closed-loop system [17]. From the acquisition stage, MMG has more advantages when compared to EMG. MMG hardware is cost effective and novel MMG sensors increases accuracy of the system [15,], good muscle force reconstruction [21] and shows good day-long recording stability [23]. MMG microphones are not sensitive to electrical and some motion artefacts, whereas EMG requires additional motion detection filters for differentiating signal and artefacts. MMG is less susceptible to electrical interference as the signal is mechanical in nature [18]. Therefore, MMG sensors are effective not only in prosthetics [22] but also in closed-loop FES.

Figure 7. Measured speed (RPM) for the full range of requested speeds.

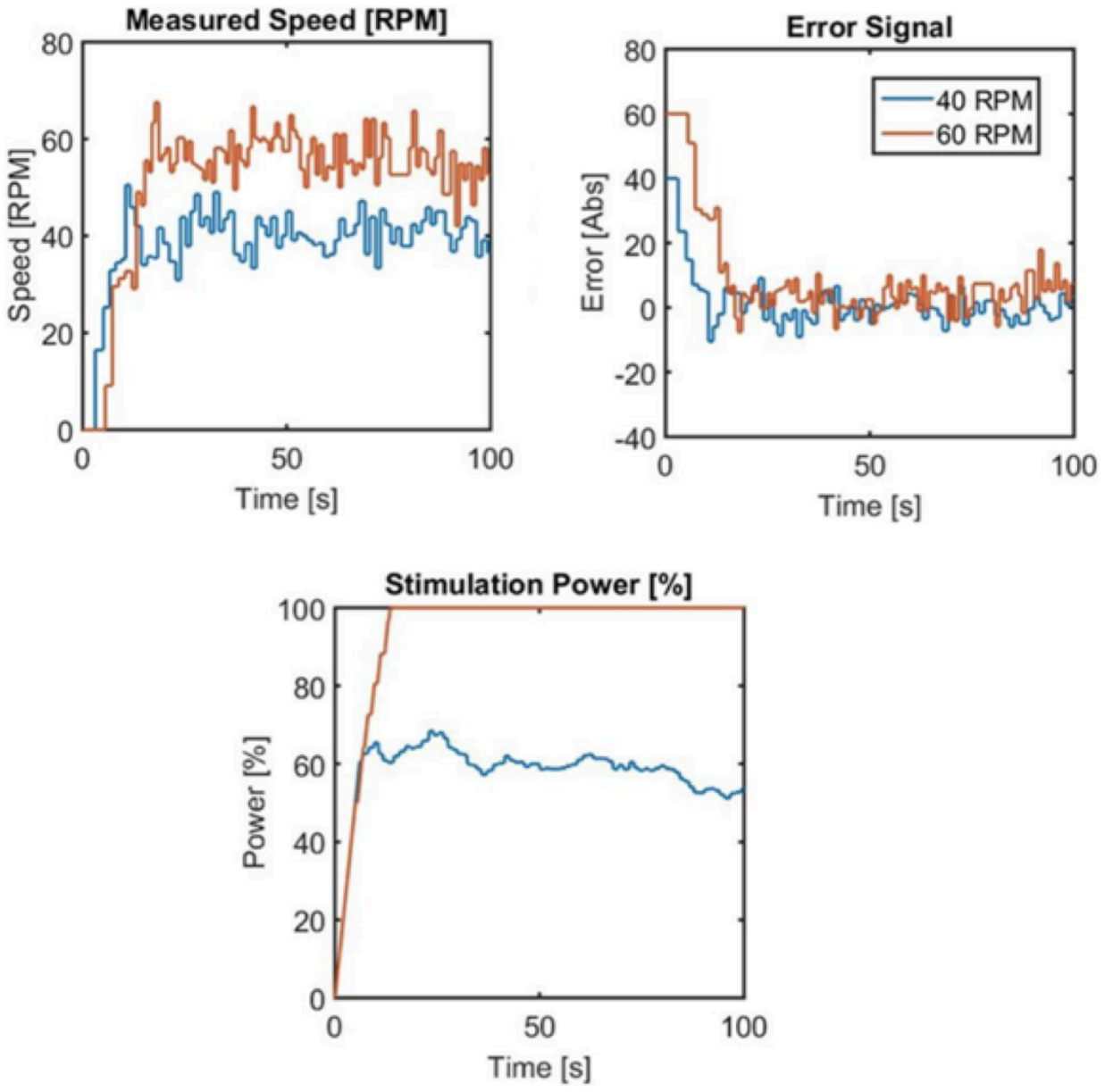

The intention for using FSRs on the pedals was that the system would have an indication of how effective the muscle contractions have been on pedal force. Unfortunately, because the crank angle encoder measurements were so problematic in Simulink the data collected from the FSRs and MMG probes were not used in the closed loop control algorithm, even though the system was able to collect it.

The development of the Simulink models was problematic. Output to the stimulator and inputs from the MMG probes, FSRs and user inputs all functioned correctly in the Simulink algorithm but there were problems with the crank angle encoder. The crank angle inaccuracies appeared to be due to limitations Simulink to process the signals in such rapid succession in real time.

The closed loop algorithm could have been successful but because of the limitations of Simulink to process the crank angle encoder in real time the error value never settled on a 0 value. A very low sampling frequency was used to attempt to correct this issue but this made the system inaccurate, so the measured angle was not always correct.

A PID controller system was developed using the blocks in Simulink but initial tests showed that the time dilation effect was increased to such a high value that it was unusable. A more sophisticated controller, such as those set out in the literature review by [13] could have been more effective, but they would have required even more processing time. So, in this case a closed loop control system was created using more simple blocks that could detect and reduce an error value. Data was collected for the crank angle encoder, indicated that the crank angle encoder was not giving stable measurements and this resulted in FES stimulations at incorrect times during cycling. A solution was not found to crank angle inaccuracy but a model was developed that reduced the error in the measured crank angle. During the evaluation process at a set speed of 60 RPM it was seen that the stimulation power needed to be raised to 100 % to reach this speed. But then at a set speed of 70 RPM the stimulation power was raised even more quickly up to 100 % but 70 RPM was not achieved. The system would need to apply more stimulation but this would have caused unacceptable discomfort during testing.

## V. CONCLUSION

The interesting part of this project was the testing and evaluation of the closed-loop system response and optimizing the Simulink models to enable smoother FES cycling. The inaccuracy of the measurements from the encoder in Simulink imposed certain limits. Nevertheless, the key part of this study was creating sensors such as MMG that could be used as forms of feedback to influence the closed-loop response. A closed loop control system was achieved, and the feedback signals were working in real time in conjunction with the rest of the system. One promising aspect of this study was that the closed loop control of the FES system was tested using real muscles and not on a simulated system [13]. Further work includes an accurate and effective way of processing all the data from the crank angle encoder during cycling. A system that records and process these signals at higher frequencies would enable much more accurate measurements of the pedal position during cycling, and the next step could be to incorporate the data from the FSRs and MMG probes. Furthermore, using a range of subjects who have paralyzed lower limbs with reduced sensitivity would also provide more accurate results. An able-bodied subject will have voluntary contractions during the cycling and the power output will be restricted by their pain tolerance during testing.